\documentclass{article}
\usepackage{ijcai22}

\usepackage{times}
\usepackage{soul}
\usepackage{url}
\usepackage[hidelinks]{hyperref}
\usepackage[utf8]{inputenc}
\usepackage[small]{caption}
\usepackage{graphicx}
\usepackage{amsmath}
\usepackage{amsthm}
\usepackage{booktabs}
\usepackage{algorithm}
\usepackage{algorithmic}
\usepackage{multirow}
\usepackage{amssymb}
\usepackage{pifont}
\usepackage{color}
\usepackage{subfigure}

\title{Channel Self-Supervision for Online Knowledge Distillation}

\author{
Shixiao Fan$^1$\and
Xuan Cheng$^1$\and
Xiaomin Wang~$^1$~\footnote{Corresponding author}\and
Chun Yang~$^1$\and\\
Pan Deng~$^1$\and
Minghui Liu~$^1$\and
Jiali Deng~$^1$\and
Ming Liu~$^1$
\affiliations
\emails
\{shixiaofan, cs\_xuancheng, chunyang, pandeng, minghuiliu\}@std.uestc.edu.cn,
julia\_d@163.com,
\{xmwang, csmliu\}@uestc.edu.cn,
}

\begin{document}

\maketitle

\begin{abstract}
Recently, researchers have shown an increased interest in the online knowledge distillation. Adopting an one-stage and end-to-end training fashion, online knowledge distillation uses aggregated intermediated predictions of multiple peer models for training. However, the absence of a powerful teacher model may result in the homogeneity problem between group peers, affecting the effectiveness of group distillation adversely. In this paper, we propose a novel online knowledge distillation method, \textbf{C}hannel \textbf{S}elf-\textbf{S}upervision for Online Knowledge Distillation (CSS), which structures diversity in terms of input, target, and network to alleviate the homogenization problem. Specifically, we construct a dual-network multi-branch structure and enhance inter-branch diversity through self-supervised learning, adopting the feature-level transformation and augmenting the corresponding labels. Meanwhile, the dual network structure has a larger space of independent parameters to resist the homogenization problem during distillation. Extensive quantitative experiments on CIFAR-100 illustrate that our method provides greater diversity than OKDDip and we also give pretty performance improvement, even over the state-of-the-art such as PCL. The results on three fine-grained datasets (StanfordDogs, StanfordCars, CUB-200-211) also show the significant generalization capability of our approach.
\end{abstract}

\section{Introduction}

In recent years, Deep neural networks have achieved great success in many computer vision tasks such as image classification~\cite{he2016deep}, object detection~\cite{lin2017focal}, semantic segmentation~\cite{long2015fully}. However, such signficant improvements often come at the expense of deeper or wider networks, hindering their development in the context of some real-world applications, for example, deployment on mobile devices. Knowledge distillation~\cite{hinton2015distilling} solves this problem by transferring the ability of high-capacity networks to compact networks.

Traditional knowledge distillation (KD) is a two-stage process. It normally requires pre-training a large teacher model beforehand and then transferring the capabilities of the teacher model to a less-parameterized smaller model. By aligning soft predictions between teachers and students, including logits~\cite{hinton2015distilling}, intermediate features~\cite{Romerofitnets2015}, activations~\cite{tung2019similarity}, etc., the student model achieves similar performance to the teacher model while significantly reducing complexity. Unfortunately, this two-stage training strategy usually entails additional training cost and pipeline complexity. 

To overcome traditional limitations, recently, online KD methods~\cite{ying2018DML,lan2018knowledge,song2018collaborative} adopted an end-to-end approach to perform collaborative and mutual learning among two or more peer networks by directly optimizing a target network. This mutual-learning strategy does promote performance improvement to some extent, but a significant problem is that as training progresses, the peer networks become more and more similar to each other due to the presence of distillation loss, producing a homogeneous phenomenon. Naturally with training, distillation among similar peer networks yields more and more limited gains, which largely affected the effectiveness of distillation~\cite{zhou2012ensemble}. Therefore, an active line of research has focused on this issue~\cite{lan2018knowledge,chen2020online,wu2021peer}. For example, OKDDip~\cite{chen2020online} hindered the homogenization among peers by equipping a diversity holding mechanism; PCL~\cite{wu2021peer} randomly enhanced the input to guarantee the discrepancy between branches.

In this paper, we propose a novel online knowledge distillation approach, \textbf{C}hannel \textbf{S}elf-\textbf{S}upervision for Online Knowledge Distillation (CSS) which introduces a self-supervised learning mechanism into knowledge distillation in hope with higher diversity and better distillation quality. Specifically, we solve the homogeneity problem among peers along three axes.

\noindent\textbf{Sample Diversity.} In this work, we introduce a feature augmentation module that stochasticly  masks out different regions of intermediate feature maps. Compared with pervious attempts that use different data augmentation on different branchs~\cite{wu2021peer}, our feature augmentation module also provides different representations of input samples, and more importantly, requires much less computational resources.

\noindent\textbf{Target Diversity.} To effetively utilize the feature-level self-supervision, we involves a label augmentation mechnasim, first coined in~\cite{lee2020self}. Specificially, a joint distribution of the original and self-supervised labels is generated and assigned to corresponding transformation. Consequently, the differences in both the representations and learning objectives in different branchs naturally facillitate the diversity between branchs and thereby enable higher performance in aggreated inference.

\noindent\textbf{Network Diversity.} As shown in~\cite{chen2020online}, the network-based settings often perform better than the branch-based settings due to the lack of enough independent parameters. Hence, we propose a dual-network multi-branch scheme which gives rise to a larger room for independent parameters to maintain the diversity between peers in hope with better distillation quality. We also include a dynamic distillation scheme using the relative utility score based on the loss of peers, allowing the peer with better utility to play a more critical role in distillation.

Extensive experiments were performed to pinpoint the efficacy of our method. Signficant and consistent improvements are observed in our approach even when compared to some state-of-the-art online knowledge distillation approachs, such as OKDDip~\cite{chen2020online} and PCL~\cite{wu2021peer}, on CIFAR-100 using different network architectures. Leveraging the feature-level supervision, the diversity of peers trained by our method is also significantly larger than other online knowledge distillation approachs, resulting in better ensemble performance. We also validated our method on different datasets (e.g. CIFAR-10, CUB-200-211, StanfordDogs, StanfordCars). The consistent performance boosts in different dataset manifests the wide applicability of our method.

The main contributions of our work are as follows:

\begin{itemize}
	\item We propose a new framework for online knowledge distillation, which constructs three diversity modules to mitigate the homogeneity problem. 
	\item We introduce a self-supervised learning mechanism based on internal pretext signals into the knowledge distillation system to strengthen the ability of network feature extraction.
	\item Extensive experiments demonstrate the effectiveness of our method, which shows better performance compared to the state-of-the-art methods.
\end{itemize}

\section{Related work}

%

\subsection{Online Knowledge Distillation}

Traditional offline knowledge distillation~\cite{hinton2015distilling,mirzadeh2020improved} compresses a high-capacity but cumbersome model into a compact one while it takes a two-stage training strategy because of its need for a sizeable pre-trained teacher, requiring more computational resources and greater time cost. Unlike traditional two-stage training, recently proposed online training strategies~\cite{ying2018DML,lan2018knowledge,chen2020online,wu2021peer} no longer require pre-trained teacher networks but learn from the predictions among multiple networks or branches. Some approaches use multiple networks to learn from each other, while others let networks share shallow modules to reduce training time and computational resources.~\cite{ying2018DML} is a typically multi-network method that learns directly from the predictions among multiple parallel networks.~\cite{lan2018knowledge,song2018collaborative,chen2020online,wu2021peer} are all multi-branch methods. It can be seen that these approaches place increasing importance on inter-network and inter-branch diversity. For example,~\cite{lan2018knowledge} generates gated ensemble logit from all branches on-the-fly as a soft target for each branch.~\cite{chen2020online} proposes a two-level distillation strategy with multiple auxiliary peers and a group leader, while utilizing an attention module to construct inter-branch diversity. To resist the homogeneity problem among branches,~\cite{wu2021peer} enhances the input $m$ times and adds Temporal mean network to each branch. 

\subsection{Self-supervised learning}

Self-supervised learning mainly uses the pretext task to mine its own supervised information from large-scale unsupervised data. In recent years, a large number of pretext tasks have been proposed in self-supervised learning, but they mostly rely on externally supervised signals, such as predicting image rotations~\cite{komodakis2018unsupervised}, image colorization~\cite{zhang2016colorful}, and patch permutation~\cite{noroozi2016unsupervised}, etc.~\cite{lee2020self} treats the self-supervised task and the original task as a joint task and constructs the joint label accordingly. The use of these external signals, although achieving good results, brings a great additional overhead.  In this paper, we adopt the transformation at the feature level proposed instead of input level as our pretext task, bringing better results and less additional overhead.

\section{Method}

\begin{figure*}[t]
	\centering
	\includegraphics[width=0.9\linewidth]{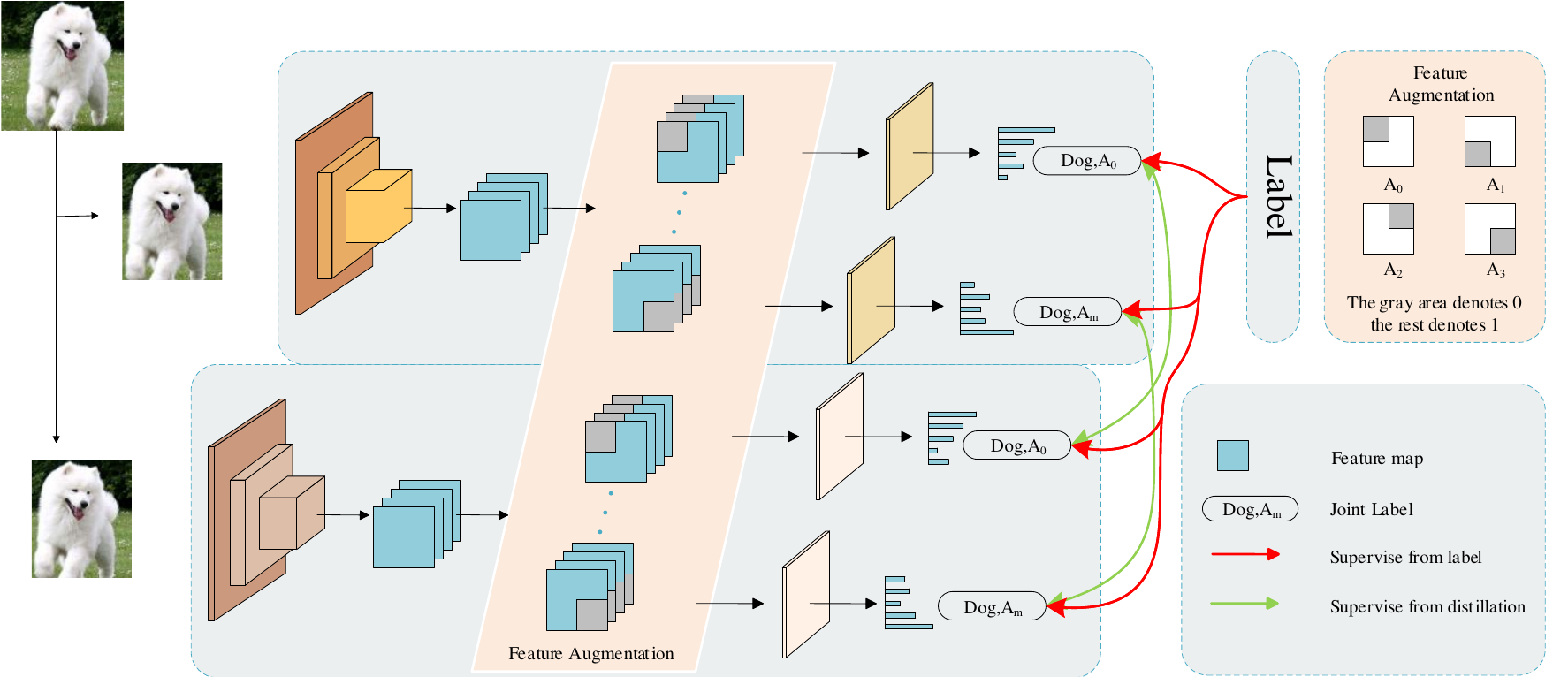}
	\caption{The overview of Channel Self-Supervision for online knowledge distillation. (Left) Three-Diversity module. Each image $x_i$ is fed into two different networks (above and below), and then the feature map ($F$) obtained from the low-level layer is augmented ($F_i$). Finally, $F_i$ is entered into the subsequent different branches. (Upper right) Schematic diagram of the feature augmentation module. (Bottom right) Explanation of some legends.}
	\label{fig:overview}
\end{figure*}

The overview of \textbf{C}hannel \textbf{S}elf-\textbf{S}upervision for Online Knowledge Distillation (CSS) is depicted in Figure \ref{fig:overview}. We use a dual-network $m$-branch structure. As shown in the Figure \ref{fig:overview}, we obtain the intermediate low-level feature map and then transform it into $m$ differentiated input for $m$ branches. Meanwhile the label space is enhanced to match the variance of the input accordingly.

\subsection{Sample Diversity}

Previous methods such as~\cite{wu2021peer} augment the sample $m$ times at input to get $m$ counterparts, and then feed each of these $m$ counterparts into $m$ branches. Obviously, this will bring $m$ times more overhead. In our work, we construct the diversity of the sample at the feature level. Like~\cite{ding2021self,yang2021self}, we extract the low-level feature map and randomly discard partial information in the same position of each channel from it to generate a variation $F'$ of the original feature $F$.
\begin{align}
	F_i = A_i \times F
\end{align}%

In practice, we usually use a simple non-overlapping mask to discard a quarter of the information in each channel of  intermediate feature maps. The significance of this is that the loss of partial information encourages the network to make better use of the global content of the image, rather than relying on a specific set of visual features. These discrepant low-level features are then fed into different branches, providing more diversity between branches whist saving considerable computional overheads compared with PCL~\cite{wu2021peer} that included different input augmentation for different peers.

\subsection{Target Diversity}

Suppose there are $M$ samples $X = \{x_i\}^M_{i=1}$ from $K$ classes, the corresponding label set are denoted as  $Y = \{y_i\}^M_{i=1}$ with $y_i \in \{1,2,...,K\}$. 

\begin{figure}
	\centering
	\begin{minipage}[b]{0.48\linewidth}
		\centerline{\includegraphics[width=1\linewidth]{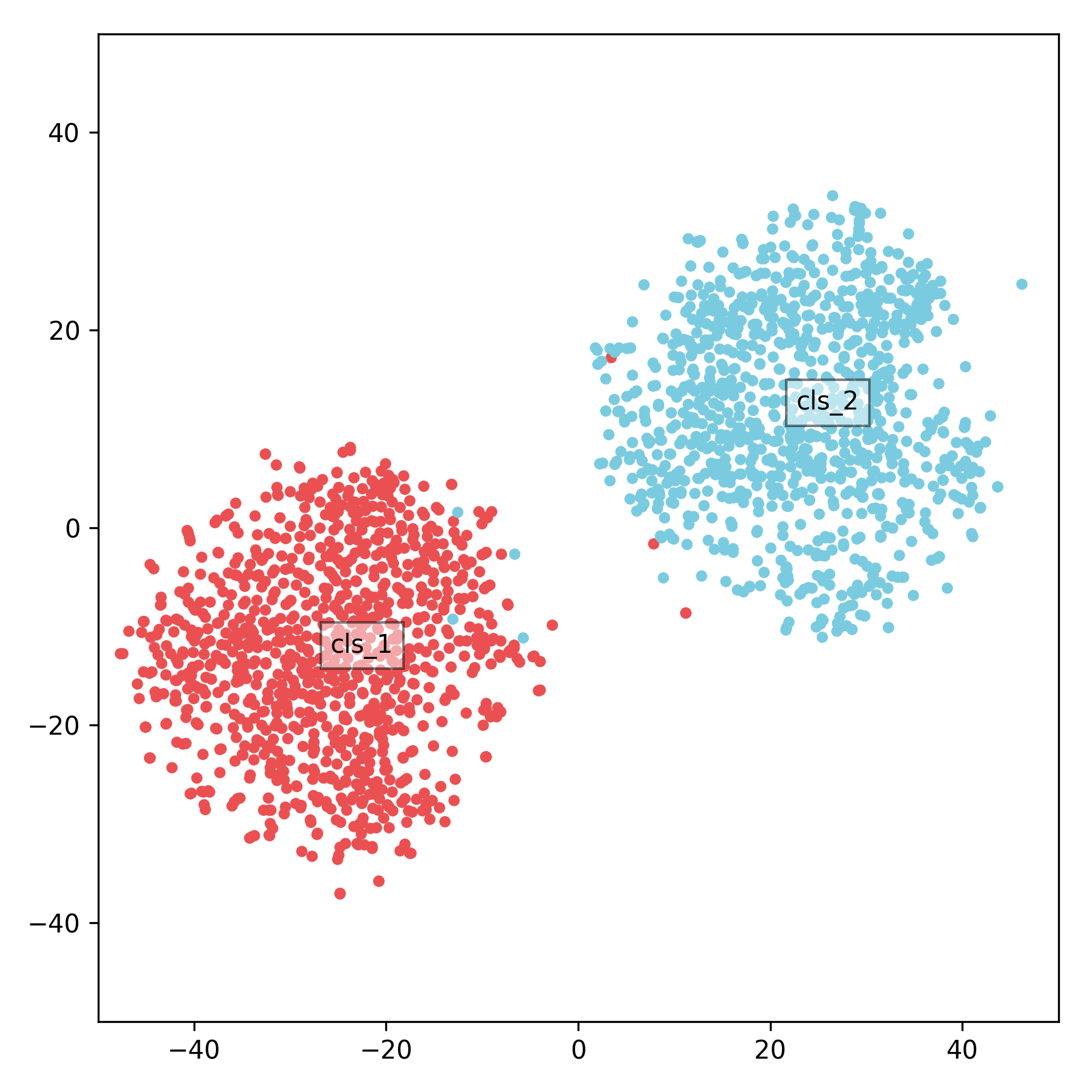}}
		\centerline{{\scriptsize (a) w/o CSS}}
	\end{minipage}
	\hfill
	\begin{minipage}[b]{0.48\linewidth}
		\centerline{\includegraphics[width=1\linewidth]{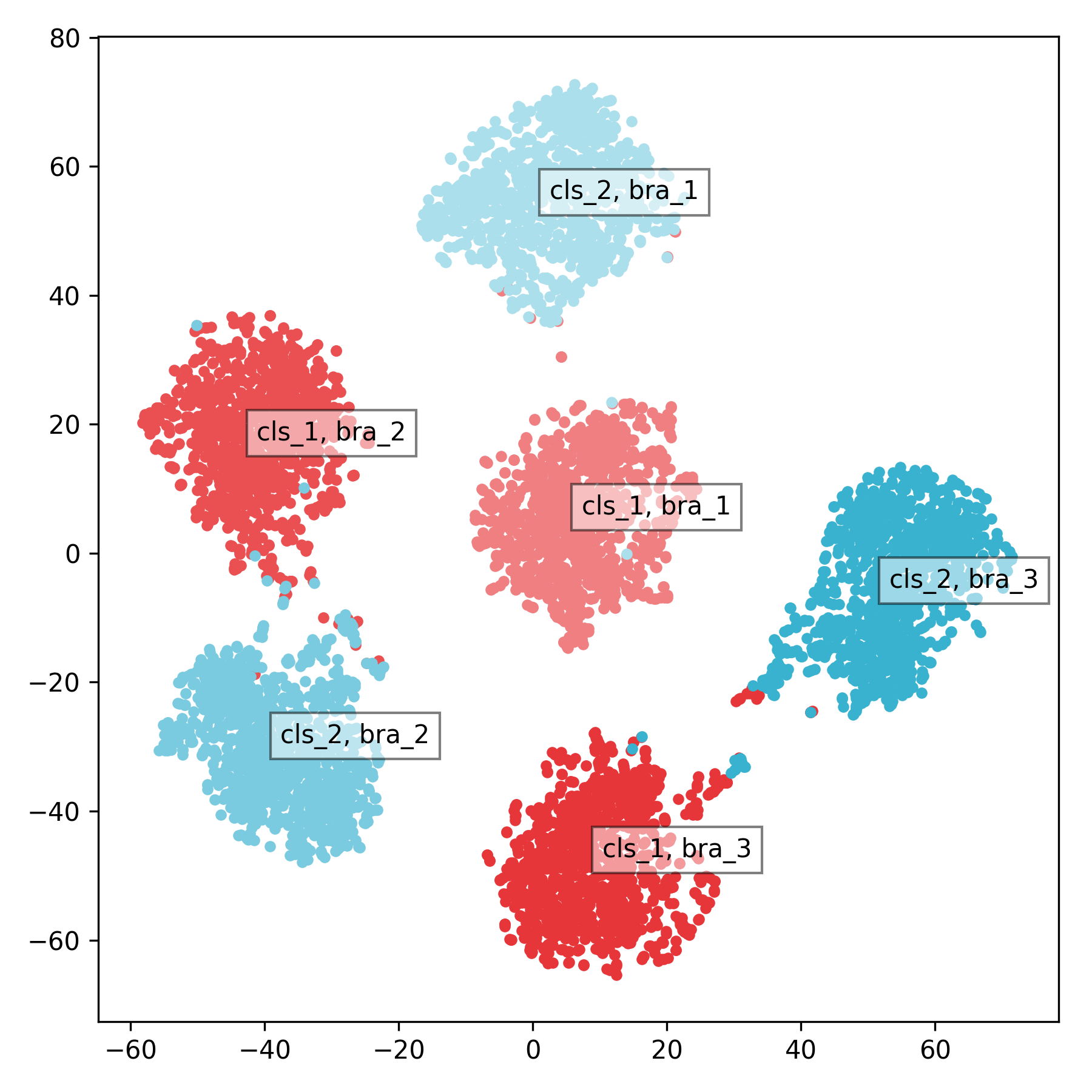}}
		\centerline{{\scriptsize (b) w/ CSS}}
	\end{minipage}
	\caption{Visualization of penultimate layer’s representations of CIFAR-10/ResNet-110 by t-SNE tool: (a) w/o CSS and (b) w/ CSS.}
	\label{fig:tsne}
\end{figure}

In self supervised learning, it is necessary to automatically generate labels to match the corresponding pretext tasks. However, previous online KD methods did not treat the labels accordingly after different enhancements of the samples. In CSS, we use a joint label to augment the original label to match the various transformations done on the feature level as~\cite{lee2020self} did. An extra label will be added to each original label to indicate different augmentations of intermediate feature maps. The joint label is denoted as  $Y = \{y_i,A_i\}$ , so that our label space is expanded to $K * m$. Unlike the original softmax classifier, here we use a joint softmax classifier $\sigma(.;\mu)$ to output the joint probability:
\begin{align}
	P(i,j|{F}') = \sigma_{ij}({F}';\mu) = \frac{exp(\mu_{ij}^T,{F}')}{\sum_{k,l}exp(\mu_{kl}^T,{F}')}
\end{align}%
The expanded target space together with the feature transformations form into a self-supervised task. Such unified task not only facilitates better representation learning(the model is enforced to recongnize different augmentations for the same sample), but also naturally encourages the difference between branches(different branches are optimized by different joint labels).  In Fig \ref{fig:tsne}, we show results of visualizing penultimate layer representations of image classifiers trained with and without our method on CIFAR-10 for classes "airplane" and "automobile". For the model trained without our method, we observe that two classes organize in two broad clusters whereas for the model with our method, each class is sucessfully spread into three different defined clusters representing three branches. The projections manifest the diversity between branches of our method.

When testing, since the exact transformations applied on each branch are already known to us, we can simply aggregate their probabilities to obtain the final probability of sample $X$:

\begin{align}
	P_{agg}(i|x) = \frac{exp(P_i)}{\sum_{k=1}^Kexp(P_k)}
\end{align}%
\begin{align}
	P_i = \frac{1}{m}\sum_{j=1}^m\mu_{ij}^T{F}'_j
\end{align}%

\subsection{Network Diversity}

One problem with the branch-based online knowledge distillation approach is that large amount of parameters are shared through branches, making the space for maintaining diversity among branches extremely compressed. For example, in~\cite{chen2020online}, network-based performance is usually better than branch-based. We believe that multiple networks are necessary in online KD because the parameters between networks are always independent providing enough space to construct diversity. In this work, we include another peer network in the distillation process. Each branch in one network is required to learn from the branch with the same transformation operation of another network. 

We also include a dynamic distillation scheme using the relative utility score based on the loss of the model. Specificially, for one network ($net1$), the distillation loss is re-weighted by the following relatively utility score:
%
\begin{align}
	\alpha_1 = \frac{L_{ce}^{net1}}{L_{ce}^{net2}} = \frac{\sum_{1}^{m}\sum_{i,j}l_{ij}log(P^{net1}_{ij}(x))}{\sum_{1}^{m}\sum_{i,j}l_{ij}log(P^{net2}_{ij}(x))}
\end{align}%
where $l_{ij} = 1$ if $y_i = j$, and 0 otherwise. Note here $j \in \{1, 2, ..., K * m\}$. For the other network, the relatively utility score would be $\alpha_2 = 1 / \alpha1$. As training proceeds, the difference between the utilities of both peer networks may cause the stronger network to be compromised by the weaker one. Our proposed method is a good solution to this, allowing the model to make a judgement at each time based on the ratio of current performances of both network and assign a higher weight to the better-optimized model. Such the dynamic distillation scheme also allows for the possibility that two different networks can be effectively distilled together, as evidenced by the experimental results (Section 4.4).

%

\begin{table*}[h]
	\centering
	\begin{tabular}{l|rrrr|rrr}
		\toprule[1.3pt]
		& DML & ONE & OKDDip & PCL & Baseline & Ours & Ensemble\\
		\midrule
		\midrule
		ResNet-56      & 76.14\small{$\pm$0.38}  & 75.94\small{$\pm$0.05}   & 76.67\small{$\pm$0.20}   & $-$  						& 74.28\small{$\pm$0.36}  & \textbf{78.20\small{$\pm$0.15}} & {78.93\small{$\pm$0.09}} \\
		ResNet-110     & 77.92\small{$\pm$0.34}  & 78.16\small{$\pm$0.19}   & 78.78\small{$\pm$0.29}   & 79.98\small{$\pm$0.55}  	& 76.21\small{$\pm$0.57}  & \textbf{80.71\small{$\pm$0.17}} & {81.56\small{$\pm$0.34}} \\
		WRN20-8        & 79.39\small{$\pm$0.13}  & 78.65\small{$\pm$0.07}   & 79.18\small{$\pm$0.13}   & 80.51\small{$\pm$0.49}  	& 78.03\small{$\pm$0.40}  & \textbf{81.12\small{$\pm$0.15}} & {81.40\small{$\pm$0.10}} \\
		ResNeXt-29     & 80.96\small{$\pm$0.20}  & 81.38\small{$\pm$0.56}   & 81.65\small{$\pm$0.28}   & 82.62\small{$\pm$0.23} 	& 79.43\small{$\pm$0.43}  & \textbf{82.83\small{$\pm$0.16}} & {83.36\small{$\pm$0.11}} \\
		PyramidNet     & 81.15\small{$\pm$0.07}  & 81.33\small{$\pm$0.39}   & 81.29\small{$\pm$0.12}   & $-$     					& 79.52\small{$\pm$0.14}  & \textbf{82.07\small{$\pm$0.13}} & {83.26\small{$\pm$0.10}} \\
		SeResNet-110   & 78.58\small{$\pm$0.02}  & 79.00\small{$\pm$0.23}   & 79.37\small{$\pm$0.26}   & $-$    					& 76.83\small{$\pm$0.15}  & \textbf{80.42\small{$\pm$0.13}} & {81.44\small{$\pm$0.10}} \\
		\bottomrule[1.3pt]
	\end{tabular}
	\caption{Top-1 accuracy (\%) with different network architectures on CIFAR-100. PyramidNet: $\alpha$ = 84, $depth$ = 110. The results of PCL are taken from its original paper.}
	\label{tab:Cifar100}
\end{table*}

Thus for dual-network multi-branch structure, the overall optimization objective consists of the following components which can be written as:
\begin{align}
	L = L_{ce}^{net1} + L_{ce}^{net2} + L_{KD}
\end{align}%
\begin{align}
\notag	L_{KD} = T^2(\alpha_1\sum_{i=1}^{m} KL(logits_i^{net1}, logits_i^{net2}) 
			    \\  + \alpha_2\sum_{i=1}^{m} KL(logits_i^{net2}, logits_i^{net1}))
\end{align}%
where KL is the Kullback-Leibler divergence and following the suggestion in~\cite{lan2018knowledge}, we set T to 3 in this paper for all methods.

\section{Experiment}

\subsection{Performance of CIFAR-100}

%


\textbf{Results} Six kind of networks are employed to verify the effectiveness of our
method, such as ResNet-56/ResNet-110~\cite{he2016deep}, WRN20-8~\cite{zagoruyko2016wide}, ResNeXt29-2$\times$64d~\cite{xie2017aggregated}, PyramidNet~\cite{han2017deep} and SeResNet-110~\cite{hu2018squeeze}. Other implementation details are present in Supplementary Materials. As shown in Table \ref{tab:Cifar100}, for all the network structures, CSS shows different degrees of advantages over vanilla settings. For example, for ResNet-110, we improved 4.5\% from the baseline, while in WRN20-8, ResNeXt29, we both outperformed more than 3\%. For the PyramidNet, our method still performs well, improving on the baseline by 2.55\%. These experimental results reveal that our approach enjoys a model-agnostic property. Besides, we also report the ensemble performance where we utilize capacities of two models. As we can see, diverse models sucessfully leads to a stronger ensemble, which not only justifies our motivation on constructing network diversity, but also reveals how powerful the generated group knowledge to some extent. For example, our method achieves an accuracy improvement of over 0.85\% after a simple average ensemble of the two networks on ResNet-110.


\textbf{Comparison with the State-of-the-Arts} Further more, We compare CSS to several recently proposed online knowledge distillation approaches, including DML~\cite{ying2018DML}, ONE~\cite{lan2018knowledge}, OKDDip~\cite{chen2020online} and PCL~\cite{wu2021peer}. Except for DML, all methods use a branch-based structure and results of all approaches are the average of the branches (OKDDip adopts the results of group-leader). We set $m$ to $3$ for all methods. We can see that our method has excellent improvement compared with the state-of-the-art. For example, on ResNet-110 and WRN20-8, our method improves 0.73\% and 0.61\%  compared with the state-of-the-art. For the more parametric networks ResNeXt29 and PyramidNet, CSS also has a significant improvement over the state-of-the-art (0.21\% and 0.74\%).

\subsection{Diversity analysis}

Next, we show the diversity that our method brings in a more intuitive way. We use the Euclidean distance between branches as a quantitative criterion for diversity, as performed in OKDDip~\cite{chen2020online} and PCL~\cite{wu2021peer}. For convenience, we set $m$ to 4 for other methods and adopted a 2-networks 2-branches structure for CSS. 

\begin{figure}[t]
	\centering
	\includegraphics[width=1.1\linewidth]{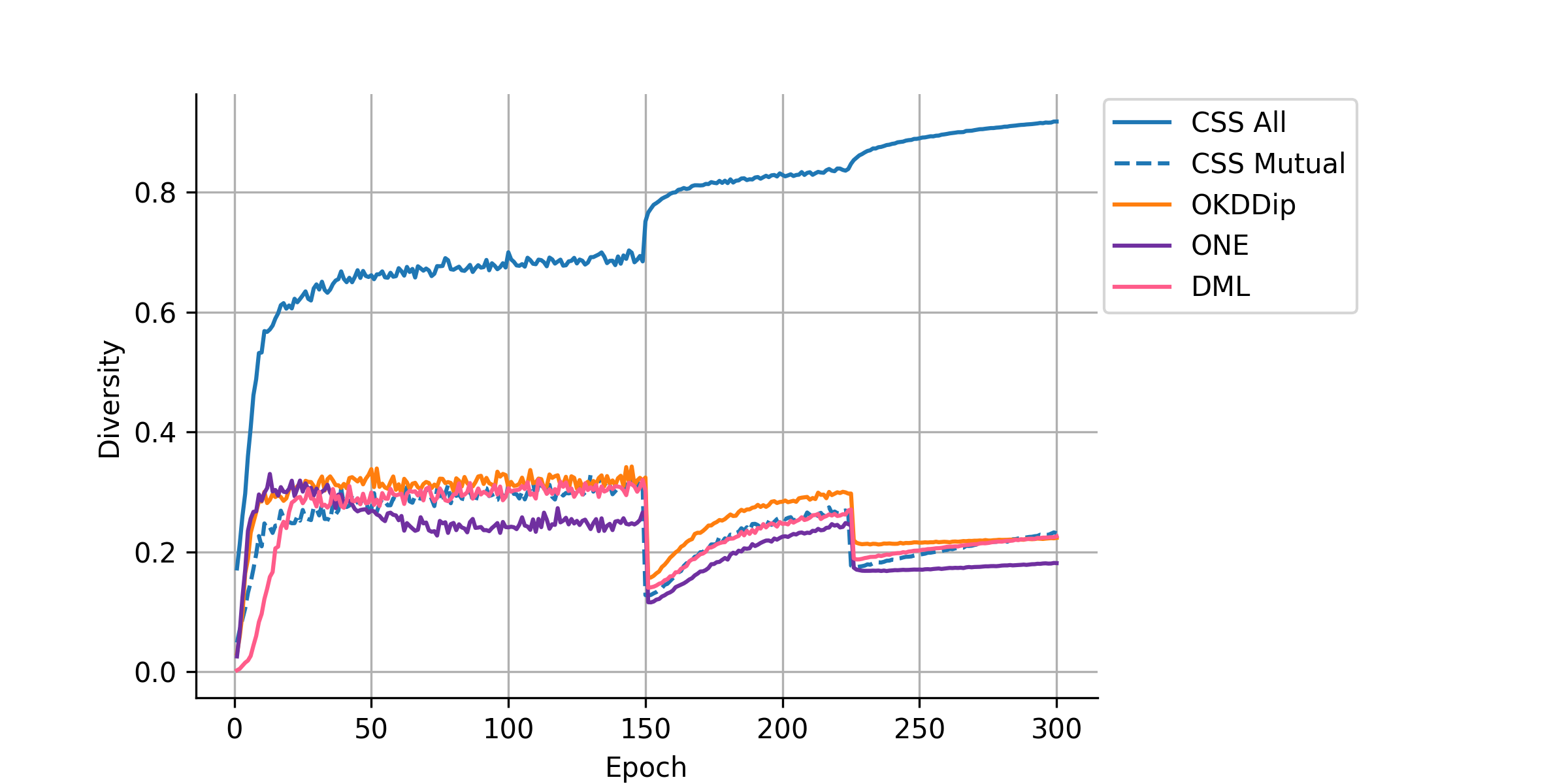}
	\caption{Branches diversity for several online KD methods with ResNet-110 on CIFAR-100. Here, we set the other methods to 4 branches and set the CSS to two networks, two branches each.}
	\label{fig:div}
\end{figure}

First, We computed the average Euclidean distance between the predictions of each pair of branches as the diversity. As depicted in the Figure \ref{fig:div}, throughout the whole epochs, the branch variance in our method (CSS) is significantly larger than other methods, demonstrating stronger diversity. For OKDDip, ONE and DML, the diversity decreases rapidly at the 150th epoch (when the learning rate changes). Then it slowly climbs up to the 225th epoch, drops again, and then stabilizes. Such fluctuations are due to the change of  the generalisation capability of branchs. In these methods, all the branchs are optimized towards the same or nearly the same objective, thereby being hard to facilitate variances of generalisation capabilities among branchs in the later stage of training. On the contrary, our method shows an increasing trend throughout the training process. This is because the self-supervised learning in our method allows different branches of the same network to handle different classification tasks. The differences in the optimized goal naturally leads to the variances of the generalization capabilities among branchs, thereby resulting in larger diversity which translates into better distillation quality and overall performance.

We then compute the diversity among the corresponding branches of different networks where we only averaged the Euclidean distances between the correponding branchs from two networks. As shown in the Figure \ref{fig:div}, our mutual diversity is comparable to DML and slightly inferior than OKDDip during the whole training process. This obversation suggests that apart from the integrated self-supervised task, the dual-network architecture still provides considerable diversity.

We also verified whether our more diverse branches have stronger integration effect than other approaches. For ONE, CL~\cite{song2018collaborative}, and CSS, we ensembled all branches, and for OKDDip, we ensembled all peers. As the Table \ref{tab:ensemble} shows, CSS achieves better results than all the other methods in both branch settings. With 4 branches, we improves 1.52\% over OKDDip and 0.87\% over ONE, and when the number of branches went to 6, we improves 1.31\%  and 0.81\% over OKDDip and ONE. It can be seen that our approach still maintains a substantial improvement in network performance when the number of branches grows.

\subsection{Performance of other datasets}

We then extended the experiments to other commonly used benchmark datasets, including CIFAR-10, Tiny-ImageNet and three fine-grained datasets (CUB~\cite{wah2011caltech}, StanfordDogs~\cite{khosla2011novel}, and StanfordCars~\cite{krause20133d}). Tiny-ImageNet is a subset of the ImageNet~\cite{russakovsky2015imagenet} dataset with 200 classes.
\begin{table}[htb]
	\centering
	\begin{tabular}{l|ccc|c}
		\toprule[1.3pt]
		Method & ONE & CL & OKDDip & Ours\\
		\midrule
		\midrule
		Top1(4)     & 80.40  & 80.05  & 79.75  & \textbf{81.27}    \\
		\midrule
		Top1(6)     & 80.75  & 80.45  & 80.25  & \textbf{81.56}    \\
		\bottomrule[1.3pt]
	\end{tabular}
	\caption{Top-1 accuracy (\%) with different network architectures of ensemble predictions: 4 branches (1st row) and 6 branches (2nd row). For CSS 4 means 2 $\times$ 2 and 6 means 2 $\times$ 3}
	\label{tab:ensemble}
\end{table}

\begin{table}[h]
	\centering
	\begin{tabular}{l|l|r|r}
		\toprule[1.3pt]
		&      				& Baseline 		& Ours 				\\
		\midrule
		\midrule
		\multirow{2 }{*}{CIFAR-10}  & ResNet-56                 	 	& 94.26\small{$\pm$0.10}                & \textbf{95.01\small{$\pm$0.06}}                     \\
		& ResNet-110                	 	& 94.42\small{$\pm$0.19}                & \textbf{95.43\small{$\pm$0.14}}                     \\
		\midrule
		Tiny-ImageNet               & ResNet-50							& 49.43\small{$\pm$0.50}				& \textbf{54.49\small{$\pm$0.73}}  					\\
		\midrule 
		CUB                       & \multirow{3}{*}{ResNet-50}    	& 53.34\small{$\pm$1.40}        		& \textbf{60.78\small{$\pm$0.88}}      				  \\
		
		StanfordDogs        &                  					    & 63.74\small{$\pm$0.16}   				& \textbf{71.59\small{$\pm$0.77}}         			  \\
		
		StanfordCars        &                  					    & 81.72\small{$\pm$1.81}   				& \textbf{87.72\small{$\pm$0.41}}					  \\
		\bottomrule[1.3pt]
	\end{tabular}
	\caption{Top-1 accuracy (\%) with different network architectures on CIFAR-10 and three commonly used fine-grained datasets.}
	\label{tab:other dataset}
\end{table}
We can see from Table \ref{tab:other dataset} that for CIFAR-10, CSS improves the baseline by approximately 1.01\% and 0.75\% with ResNet-110 and ResNet-56, respectively. For Tiny, CSS gets the improvement over baseline almost 5.06\%. For fine-grained classification tasks, there are only subtle differences between samples, so the model must pay more attention to the details of the samples. We adopt ResNet-50 to demonstrate the performance of CSS on fine-grained datasets to confirm the generalizability of our approach to fine-grained classification tasks. As shown in Table \ref{tab:other dataset}, CSS has significant improvements on three fine-grained datasets. For CUB and StanfordCars, our method improves over the baseline by 7.44\% and 6\%, respectively, while for StanfordDogs, CSS improves by almost 7.85\%. These results indicate that our proposed CSS is dataset-agnostic.

\subsection{Analysis and Discussion}

\begin{table*}[h]
	\centering
	\begin{tabular}{c|l|c|c|c}
		\toprule[1.3pt]
		\multicolumn{1}{c}{\multirow{2}{*}{Net1}}     & \multicolumn{1}{c}{\multirow{2}{*}{Net2}} &\multicolumn{2}{c}{Top-1 (\%)} &  \multicolumn{1}{c}{\multirow{2}{*}{Ensemble}}\\
		\cmidrule(lr){3-4}
		\multicolumn{1}{c}{}                     	  & \multicolumn{1}{c}{}                      & \multicolumn{1}{c}{Net1}      & \multicolumn{1}{c}{Net2} & \multicolumn{1}{c}{}                          \\ \cmidrule(r){1-5} 
		\multirow{5}{*}{ResNet-110} 
		& ResNet-56             &   80.29\small{$\pm$0.12} (\textcolor{green}{4.08 $\uparrow$})    &     78.70\small{$\pm$0.16} (\textcolor{green}{4.42 $\uparrow$}) & 80.54\small{$\pm$0.10}               \\
		& ResNet-110            &   80.65\small{$\pm$0.16} (\textcolor{green}{4.44 $\uparrow$})    &     80.76\small{$\pm$0.21} (\textcolor{green}{4.55 $\uparrow$}) & 81.56\small{$\pm$0.15}               \\
		& SeResNet-110          &   80.68\small{$\pm$0.12} (\textcolor{green}{4.47 $\uparrow$})    &     80.56\small{$\pm$0.07} (\textcolor{green}{3.73 $\uparrow$}) & 81.82\small{$\pm$0.24}               \\
		& PyramidNet            &   81.02\small{$\pm$0.13} (\textcolor{green}{4.81 $\uparrow$})    &     82.90\small{$\pm$0.23} (\textcolor{green}{3.38 $\uparrow$}) & 83.33\small{$\pm$0.36}				\\
		\bottomrule[1.3pt]                
	\end{tabular}
	\caption{Top-1 accuracy (\%) on different networks distill each other. The green content is the change relative to the baseline, and the upward arrow indicates elevation.}
	\label{tab:diff}
\end{table*}

\begin{figure}[t]
	\centering
	\includegraphics[width=0.8\linewidth]{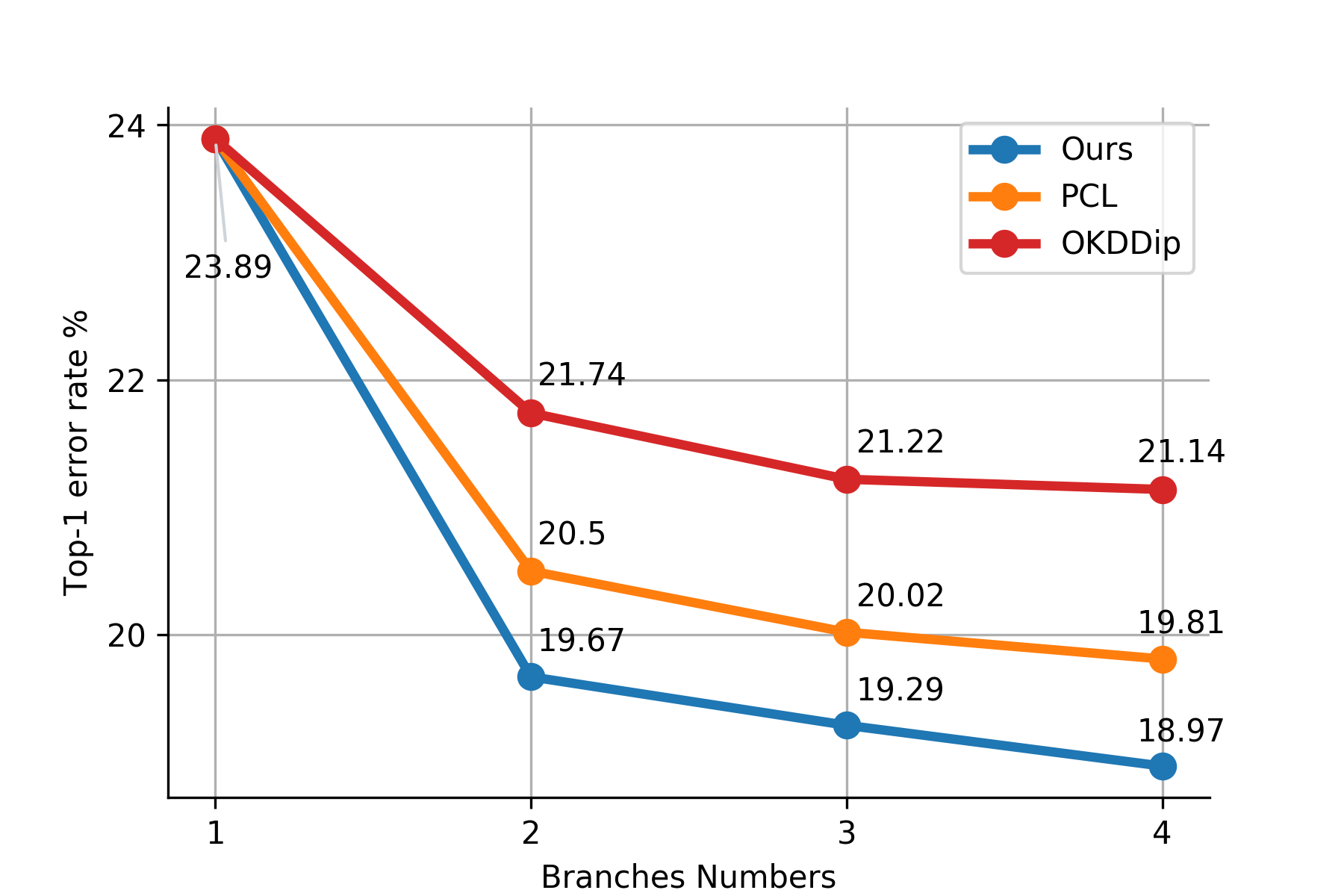}
	\caption{Top-1 error rates (\%) on different branch number. The results of PCL are taken from its original text. }
	\label{fig:branch}
\end{figure}

\noindent\textbf{Branches Number.} We evaluated the impact of different branch numbers on performance. The Figure \ref{fig:branch} shows that we compare ours with PCL and OKDDip on CIFAR-100 using ResNet-110. As the number of branches increases, we consistently perform better than other methods. We believe this is because CSS constructs a more effective diversity than other methods, which nicely moderates the homogenization problem, so more branches give better results for distillation. On top of that, we also have more boosts. The CSS improved by almost 0.32\% (from 19.29 to 18.97) and the PCL improved by 0.21\% (from 20.02 to 19.81 as reported in~\cite{wu2021peer}) when the number of branches was raised from 3 to 4.

\noindent\textbf{Distillation with different networks.} As a dual-network multi-branch framework, our method allows for the possibility that two different networks can be distilled together. In light of this, we evaluated the performance of our method on distillation between ResNet-110 and other networks. As shown in Table \ref{tab:diff},  we can see that: (1) Both networks in our framework obtain consistent and signficant performance boosts over their vanilla settings. (2) Our method can still lead to a better ensemble performance even when capacities of two networks differ by a large margin. This could attribute to both the diversity construction mechanism that allows the weaker network to provide effective distillation instruction, and the dynamic distillation scheme that guarantees the stronger network to play a morecritical role when training. (3) Another interesting point is that when distilled with different networks, the performance gains are not as large as the differences in capacities of networks. When distilled with ResNet-56, ResNet-110 and PyramidNet, ResNet-110 achieves 80.29\%, 80.65\% (0.26\% over ResNet-56 counterpart) and 81.02\% (0.73\% over ResNet-56 counterpart), respectively. The fact that the model could still achieve comparable and considerable perfomance gain even when distilled with weaker networks not only provides another proof of the success and efficiency of diversity construction mechanism, but also allows us to further reduce the overhead in some resource-constrained scenarios.

\begin{table}[h]
	\centering
	\begin{tabular}{c|llll|c}
		\toprule[1.3pt]
		&SD & TD & ND & DW & CIFAR-100\\
		\midrule
		\midrule
		Backbone    &           &            &            &            & 76.21 			\\
		\textbf{A}	&\ding{51}	&   	     &   		  &   		   & 79.88    		\\
		\textbf{B}	&\ding{51}	& \ding{51}  &   		  &   		   & 80.15    		\\
		\textbf{C}	&\ding{51}	& \ding{51}  & \ding{51}  &   	 	   & 80.97    		\\
		\textbf{D}	&\ding{51}	& \ding{51}  & \ding{51}  & \ding{51}  & \textbf{81.27} \\
		\bottomrule[1.3pt]
	\end{tabular}
	\caption{Ablation study: Top-1 accuracy (\%) for ResNet-110 on CIFAR-100. For \textbf{A} and \textbf{B}, we adopt 1 $\times$ 4 (1 means one network, 4 means four branches). \textbf{C} and \textbf{D} : 2 $\times$ 2.}
	\label{tab:ablation}
\end{table}

\subsection{Ablation Study}

In this section, we study the effect of different elements in CSS. We chose ResNet-110 as our baseline setting and applied dual-network 2-branch CSS to perform the ablation experiments. The results reported are those obtained from the ensemble of different branches.

\noindent\textbf{The effect of Sample Diversity.} Case A denotes the network, which adds \textbf{S}ample \textbf{D}iversity (SD) to the backbone. It can be seen that Case A is over 3 percentage points higher than the baseline.

\noindent\textbf{The effect of Target Diversity.} Case B is based on Case A with the corresponding enhancements to the label. As the table \ref{tab:ablation} shows, Although Case A has achieved a high level, Target Diversity module still improves the performance of the network (+0.27\%). This shows that the network learned more after TD differentiated the optimization objectives of different branches.

\noindent\textbf{The effect of Network Diversity.} Case C demonstrates the advantages of the dual network schema we have adopted. It can be seen that the performance is improved by 0.82\% after taking dual network on top of Case B. Specifically, here we have used a fixed factor for distillation losses.

\noindent\textbf{The effect of Dynamic Weight.} To evaluate the effect of dynamic weights, we also conduct Case D. As the results in the last two lines in Table \ref{tab:ablation} show, the use of dynamic weights improves the classification accuracy of the network by 0.30\%. Because of the boost gained from 80.97\%, we believe the presence of DW is crucial. The results in Section4.4 also provide a good illustration of where the advantages of dynamic weights can be found. When ResNet-110 and different networks are distilled from each other, setting a fixed weight for these network models with capacity differences shows that it is not a smart idea.

\section{Conclusion}

In this paper, we propose a novel online knowledge distillation framework (CSS), which adopts a dual-network multi-branch structure and alleviates the homogeneity problem from samples, targets, and networks. By implementing the feature-level transformation and augmenting the corresponding labels, we improve inter-branch diversity through this self-supervision approach. Extensive experiments show that CSS constructs better diversity even has better performance over the state-of-the-art on CIFAR-100. The results on CIFAR-10 and three fine-grained datasets also indicated the great generality of our approach.

\bibliographystyle{named}
\bibliography{ijcai22}

\end{document}